\documentclass[letterpaper]{article} 
\usepackage{aaai24}  
\usepackage{times}  
\usepackage{helvet}  
\usepackage{courier}  
\usepackage[hyphens]{url}  
\usepackage{graphicx} 
\usepackage{natbib}  
\usepackage{caption} 
\usepackage{amsmath,amssymb,amsfonts}
\usepackage{graphicx}
\usepackage{textcomp}
\usepackage{algorithm2e}
\RestyleAlgo{ruled}
\SetKwComment{Comment}{/* }{ */}
\usepackage{multirow}
\usepackage{colortbl}
\usepackage{booktabs}
\usepackage{tabulary}
\usepackage{xcolor}
\usepackage{pifont}
\usepackage{diagbox}
\usepackage{vcell}
\usepackage{array}
\usepackage{xfrac} 
\usepackage{nicefrac}
\newcolumntype{C}[1]{>{\centering\arraybackslash}m{#1}}
\newcolumntype{?}[1]{!{\vrule width #1}}

\def\ie{\emph{i.e.}}
\def\eg{\emph{e.g.}}
\def\etal{{\em et al.}}

\title{Memory-Efficient Prompt Tuning for Incremental Histopathology Classification}
\author{
    Yu Zhu\textsuperscript{\rm 1,2}\equalcontrib,
    Kang Li\textsuperscript{\rm 1}\equalcontrib\thanks{Corresponding Author.},
    Lequan Yu\textsuperscript{\rm 3},
    Pheng-Ann Heng\textsuperscript{\rm 1},
}
\affiliations{
    \textsuperscript{\rm 1}Department of Computer Science and Engineering, The Chinese University of Hong Kong\\
    \textsuperscript{\rm 3}Department of Statistics and Actuarial Science, The University of Hong Kong\\
    \textsuperscript{\rm 2}Department of Mechanical Engineering, The University of Hong Kong\\
    \{yzhu, kli, pheng\}@cse.cuhk.edu.hk, lqyu@hku.hk

}

\begin{document}

\maketitle
\begin{abstract}
Recent studies have made remarkable progress in histopathology classification.
Based on current successes, contemporary works proposed to further upgrade the model towards a more generalizable and robust direction through incrementally learning from the sequentially delivered domains.
Unlike previous parameter isolation based approaches that usually demand massive computation resources during model updating, we present a memory-efficient prompt tuning framework to cultivate model generalization potential in economical memory cost.
For each incoming domain, we reuse the existing parameters of the initial classification model and attach lightweight trainable prompts into it for customized tuning.
Considering the domain heterogeneity, we perform decoupled prompt tuning, where we adopt a domain-specific prompt for each domain to independently investigate its distinctive characteristics, and one domain-invariant prompt shared across all domains to continually explore the common content embedding throughout time.
All domain-specific prompts will be appended to the prompt bank and isolated from further changes to prevent forgetting the distinctive features of early-seen domains.
While the domain-invariant prompt will be passed on and iteratively evolve by style-augmented prompt refining to improve model generalization capability over time.
In specific, we construct a graph with existing prompts and build a style-augmented graph attention network to guide the domain-invariant prompt exploring the overlapped latent embedding among all delivered domains for more domain-generic representations.
We have extensively evaluated our framework with two histopathology tasks, \ie, breast cancer metastasis classification and epithelium-stroma tissue classification, where our approach yielded superior performance and memory efficiency over the competing methods.
\end{abstract}
%
%
%
\section{Introduction}

Histopathology classification is a fundamental task in cancer diagnosis. It aims to specify the malignancy and benignity of suspected tissues by microscopic examination.
The resulting analysis is normally considered the gold standard in determining the presence and spread of certain cancers~\cite {bejnordi2017diagnostic}.
Although recent deep-learning models have achieved remarkable progress on this task, contemporary studies are not content with the achievements made so far but strive to upgrade and update model functionality toward perfection by incremental learning~\cite{derakhshani2022lifelonger, li2022domain}.

One practical yet challenging direction for model upgrading is to incrementally boost its generalization potential over heterogeneous histopathology data.
Depending on the technician skills and digital scanner brands in different medical centers, the histology data sampled from multiple sites (\ie, domain) often exhibit heterogenous appearances after hematoxylin and eosin (H\&E) staining, varying from dark blueish purple to light pinkish purple~\cite{lin2019fast}.
Then, domain incremental learning (DIL), \ie, a model updating paradigm that enables the model to progressively adapt to more and more heterogenous domains as time goes by, would be substantial for robust histopathology classification.
For any updated model, the basic requirement is to keep the existing capability unaffected, \ie, not catastrophically forgetting the previously-acquired domains.
Moreover, we expect to enhance its generalization ability, \ie, not only well adapted to the currently delivered domains but also the unseen domains that might be encountered in the future.
Particularly in the medical field, for each update, the model would have no access to the early-delivered domains due to data privacy concerns~\cite{li2021fedbn} and storage burden~\cite{lin2019fast}.
In addition, the domain identity, \ie, the label indicating which domain one particular sample comes from, is erased as part of patient privacy during data anonymization and would be unavailable to use for model training and testing during the entire learning lifespan~\cite{gonzalez2020wrong}.

The straightforward approach is to finetune the previous model with each sequentially incoming domain one by one.
However, with the absence of past domains and data heterogeneity throughout time, it inevitably overrides and disrupts the parameters learned for past domains, leading to catastrophic forgetting of them~\cite{li2017learning}.
A promising way for this issue is to address it from the model-centric perspective, \ie, isolating the early-acquired parameters (\eg, the whole model) into separate storage and allocating new parameters to acquire the newly-arrived domain~\cite{gonzalez2020wrong, miao2022continual,li2019learn}.
Despite their effectiveness, most of them are extremely memory-intensive with increasing computation demands and memory usage over time, greatly limiting their applicability in gigapixel-sized histopathology images (\eg, 1-3 GB per slide~\cite{zhao2019pfa}).

Fortunately, we could borrow some insights from recent advances in prompt-based natural language processing approaches~\cite{su2022transferability}, which pointed out that employing learnable prompt tokens as the parameterized inputs, could encode necessary guidance to conditionally adapt the frozen pre-trained model to the downstream target task.
Inspired by that, it would be unnecessary to completely adjust the previous model to accommodate the currently delivered domain, nor isolate the entire model parameters in separate memory units to retain early-acquired knowledge.
Alternatively, it could be more memory-efficient to simply perform prompt tuning upon the initial well-trained classification model and save these lightweight prompts for future usage instead.

In this paper, we present a memory-efficient prompt tuning framework to incrementally learn from the sequentially-delivered heterogeneous domains, progressively cultivating the histopathology classification model towards a more generalizable and robust direction over time.
Considering the data heterogeneity of the domains delivered in different time steps, we perform decoupled prompt tuning with two types of prompts.
We employ a domain-specific prompt for each domain to independently investigate its distinctive features while maintaining a domain-invariant prompt shared across all domains to continually explore the common content embedding over time.
For each incoming domain, we freeze the initial model and train two lightweight prompts upon the existing weights for memory and computation efficiency.
We learn a domain-specific prompt from scratch and learn the shared domain-invariant prompt iteratively upon the previous one via style-augmented prompt refining.
Specifically, we build up a graph with the existing prompts and constrain the domain-invariant prompt to explore the co-existing and domain-agnostic representations among all seen domains via graph attention propagation. 
Meanwhile, we augment the style variations met in the prompt refining process to expose more domain-generic representations and further boost its generalization potential.
At the end of each time step, we store all prompts in the bank under economical memory costs. 
The domain-specific prompts would be isolated from further changes and retrieved later to prevent forgetting early-seen domains.
The domain-invariant prompt would be carried forward to incrementally acquire more domain-generic features to improve generalization ability.
We have extensively evaluated our framework on two histopathology classification tasks, including breast cancer metastase classification on the Camelyon17 dataset~\cite{bandi2018detection} and epithelium-stroma tissue classification on a multi-site data collection.
In both tasks, our approach showed superior performance over competing methods with better generalization on unseen domains and less forgetting of past domains.
Our main contributions could be summarized as follows:
\begin{itemize}

    \item We proposed a memory-efficient prompt tuning framework to iteratively upgrade the model towards a more generalized direction in economical memory cost.

    \item We performed decoupled prompt tuning with a series of domain-specific prompts and a shared domain-invariant prompt to tackle the heterogeneity of incoming domains.

    \item We presented style-augmented prompt refining to iteratively evolve the domain-invariant prompt over time to boost its generalization potential on unseen data.

    \item We have validated our approach on two histopathology image classification tasks, where our framework outperformed other comparison methods significantly.
\end{itemize}
%
%
%

\begin{figure*}[t]
  \centering
  \includegraphics[width=0.85\textwidth]{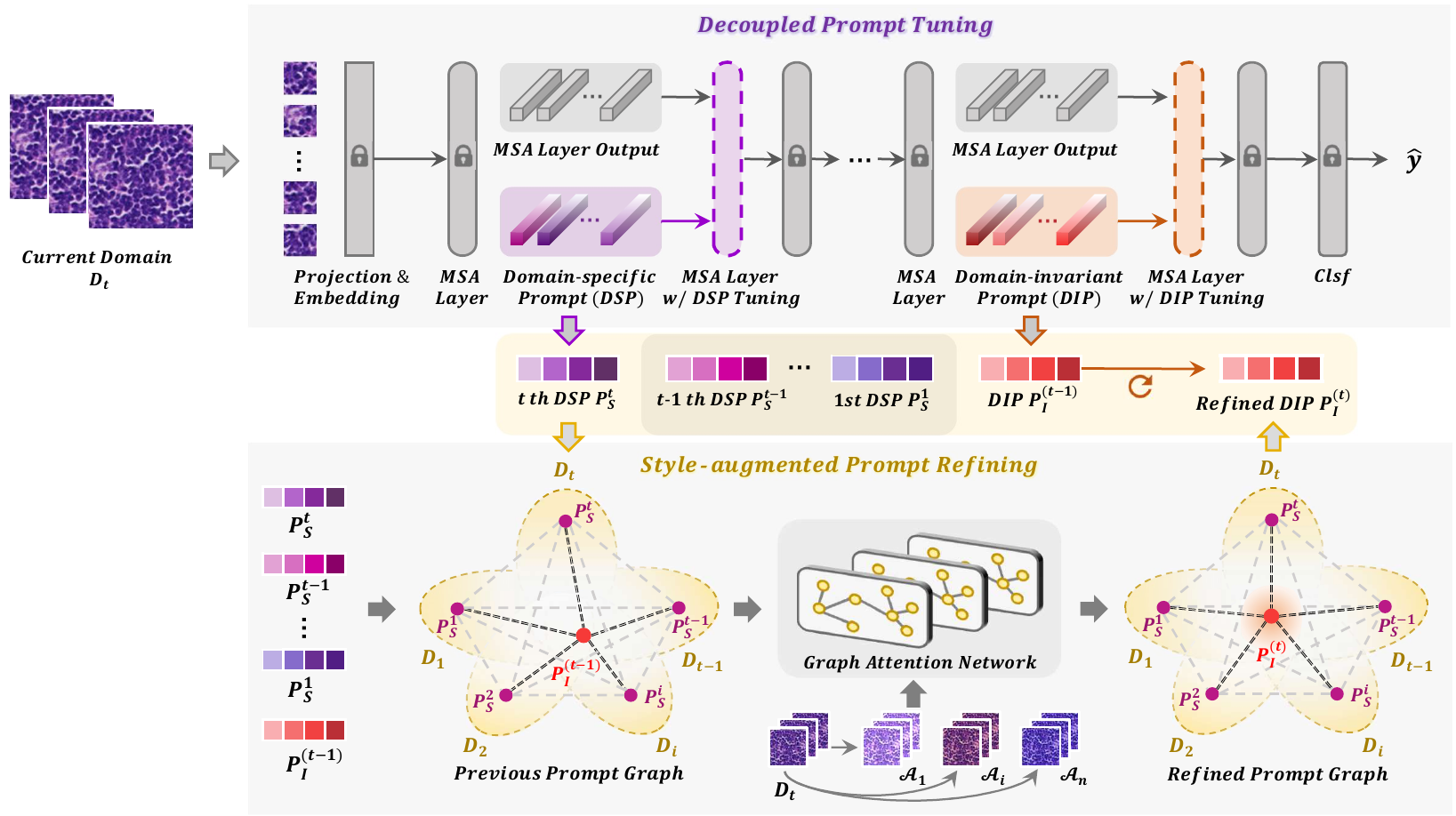}
  \caption{Overview of our memory-efficient prompt tuning framework. 
  We proposed to perform decoupled prompt tuning upon the initial model with two lightweight prompts, aiming to acquire the latest domain knowledge in economical memory cost.
  We employ a domain-specific prompt independently for each domain to acquire its distinctive features like appearances.
  The learned domain-specific prompt would be stored and isolated in the prompt bank to help alleviate the forgetting of early-acquired domains.
  Meanwhile, we maintain a domain-invariant prompt shared across domains to progressively learn the common content over time like shape prior.
  We performed style-augmented prompt refining upon the previous domain-invariant prompt, where we constrain its exploration scope within the overlapped latent embeddings of all seen domains and guide it to learn the domain-generic representations, gradually strengthening the generalization potential over time.
  }
  \label{fig:main-method}
\end{figure*}
%
%
%
\section{Related Work}
\subsection{Domain Incremental Learning}

Considerable efforts have been devoted to domain incremental learning to progressively cultivate the model accommodating more and more heterogeneous domains.
One stream of works would not require any additional module to support model updating~\cite{aljundi2018memory, li2017learning, kirkpatrick2017overcoming, zenke2017continual}.
For example, the regularization-based methods~\cite{kirkpatrick2017overcoming, aljundi2018memory} employed a loss term to penalize large changes of the parameters important to historical domains to help retain early-acquired knowledge.
However, these approaches often suffered from interval forgetting when dealing with a long sequence of incremental learning tasks, and their performance still has certain improvement spaces~\cite{luo2020appraisal, mai2022online}.
Other streams of work (\eg, replayed-based methods and parameter isolation methods) sacrificed memory usage to trade for better model performance.
For example, Shin~\etal~\cite{shin2017continual} employed an extra generative adversarial network (GAN) ($\approx$ 266MB) to memorize and replay past domain distributions to prevent forgetting past domains, while Gonzalez~\etal~\cite{gonzalez2020wrong} stored all previously-learned models ($\approx$ 81MB each) in a separate space and maintain an autoencoder-based domain classifier to retrieve them back when necessary.
Although the above methods could effectively alleviate the forgetting of historical domains, it also results in massive memory consumption, making them less applicable to gigabyte-size histopathology images.
In contrast, we maximally reuse the existing initial model and perform decoupled prompt tuning upon it by two lightweight prompts ($\approx$ 0.5MB) for each incoming domain, greatly boosting memory efficiency.

\subsection{Prompt Learning}

Inspired by the recent progress of prompt tuning in natural language processing~\cite{su2022transferability}, contemporary works attempted to apply it for incremental learning.
Most prior works concentrated on the class incremental learning (CIL) settings, \ie, progressively learning to categorize more and more classes over time.
They prevented forgetting early-acquired classes by creating a shared prompt pool for instance-wise prompt query~\cite{wang2022learning}, setting general prompts and expert prompts to form complementary learning~\cite{wang2022dualprompt} and \textit{etc}.
However, most of them are less prepared for the domain incremental learning settings, especially the demand to generalize to unseen data.
Most CIL approaches would not expect the model to correctly recognize the objects of unseen classes (\ie, never learned in training).
However, in domain incremental learning settings, it is highly desired for a model well generalizing to unseen domains of unknown appearances for robust classification.
Very recently, a DIL approach S-Prompt~\cite{wang2022s} tried to independently learn the prompts across domains for a win-win game but still overlooked the generalization issue.
In our work, we put extra effort to maintain a domain-variant prompt shared over time by style-augmented prompt refining, incrementally absorbing more domain-generic features to improve generalization.
%
%
%
%
\section{Methodology}

In domain incremental learning (DIL) settings, we assume a heterogenous data stream $D_{1}, D_{2},..., D_{T}$ sequentially delivered from multiple sites one by one.
With the arrival of the dataset $D_t$ at time step $t$, 
our goal is to incrementally optimize the previous model $M_{t-1}$ with $D_t$, such that the updated model $M_t$ would not catastrophically forget past domains $D_{1}, D_{2},..., D_{t-1}$, while maintaining satisfying generalization ability for unseen domains.
For privacy concerns in medical fields, all past domains would be inaccessible and no domain identity would be available.

Fig.~\ref{fig:main-method} overviews our framework.
For each incoming domain, we reuse the initial model and perform decoupled prompt tuning upon it with two lightweight prompts to acquire new domain knowledge in a memory-efficient manner.
In specific, the domain-specific prompt (DSP) is independently learned from scratch to tackle the distinctive features, while the domain-invariant prompt (DIP) is iteratively evolved from the previous one by style-augmented prompt refining to incrementally explore domain-generic features.

\subsection{Decoupled Prompt Tuning}\label{Sec_DPT}
We construct a transformer backbone (\eg, ViT~\cite{dosovitskiy2020image}) for the classification model.
It consists of a basic transformer feature extractor $f_b$ to convert the input image into sequence-like high-level representations, and a classification layer $f_\phi$ to map the representation to the final prediction $\hat{y}$.
At time step $t$, with the arrival of the current domain $D_t$, we load the pre-trained weights into the basic feature extractor following prior works~\cite{wang2022dualprompt,wang2022learning} and freeze them.
Upon it, we perform decoupled prompt tuning by two lightweight trainable prompts, \ie, one domain-invariant prompt $p_I^{(t)}$ and one domain-specific prompt $p_s^{t}$, to acquire the current domain.
To avoid any confusion, we use the superscript $(t)$ to denote the shared domain-invariant prompt learned in the $t$-th time step, while using the superscript $t$ to indicate the $t$-th domain-specific prompt.

The domain-invariant prompt and the domain-specific prompt can be inserted as additional inputs of any multi-head self-attention (MSA) layer in the basic transformer feature extractor.
Take the ${i}$-th MSA layer as an example.
Before passing the previous MSA layer outputs $h_{i-1} \in \mathbb{R}^{l\times m}$ to it, we keep the query $h_{i-1}^q$ and append the domain-invariant prompt $p_I^{(t)}\in \mathbb{R}^{l\times m}$ in its key $h_{i-1}^k$ and value $h_{i-1}^v$ to guide it explore the domain-shared representations as
\begin{equation}
    h_{i} = f_{MSA}^{(i)}(h_{i-1}^q, [p_I^k; h_{i-1}^k], [p_I^v; h_{i-1}^v]),
\end{equation}
where $f_{MSA}^{(i)}$ and $h_{i}$ denote the ${i}$-th MSA layer and its output tuned with the domain-invariant prompt respectively. 
$p_I^k\in \mathbb{R}^{l/2 \times m}$ and $p_I^v\in \mathbb{R}^{l/2 \times m}$ are split from $p_I^{(t)}$ to maintain the same sequence length before and after the MSA layer.
The domain-specific prompt $p_s^t \in \mathbb{R}^{l\times m}$ can be attached in a similar way to learn the distinctive features as 
\begin{equation}
    h_{j} = f_{MSA}^{(j)}(h_{j-1}^q, [p_s^k; h_{j-1}^k], [p_s^v; h_{j-1}^v]),
\end{equation}
where $f_{MSA}^{(j)}$ and $h_{j}$ denote the ${j}$-th MSA layer and its outputs respectively. 
$p_s^k, p_s^v \in \mathbb{R}^{l/2 \times m}$ are split from $p_s^t$.

As domain identity is not available during inference, we additionally equip a distinguishable key value $k_t$ for the domain-specific prompt, to help pair each test image with a matching domain-specific prompt.
We decompose each image $x_i \in D_t$ into the amplitude spectrum $\mathcal{A}{(x_i)}$ and phase spectrum $\mathcal{C}{(x_i)}$ in the frequency space by fast Fourier transform $\Phi_{\text{FFT}}$.
Since the amplitude captures the low-level statistics (\eg, style, appearances) while the phase extracts the high-level features (\eg, content, shape)~\cite{jiang2022harmofl,liu2021feddg}, we implement the key value $k_t$ as the average amplitude spectrum of the images in $D_t$ as 
\begin{equation} 
\begin{aligned}
    k_t=\frac{1}{N_t}\sum_{i=1}^{N_t} \mathcal{A}{(x_i)},
\end{aligned}
\label{eq: dsp-t-key}
\end{equation}
where $N_t$ denotes the total number of training data in $D_t$.

In the first time step ($t=1$), we simultaneously optimize the domain-specific prompt $p_s^1$, the domain-invariant prompt $p_I^{(1)}$ and the classification layer $f_{\phi}$ upon the frozen basic feature extractor $f_b$ by the training samples $(x,y) \in D_1$ as
\begin{equation} 
\min_{f_{\phi}, p_I^{(1)}, p_s^1} \mathcal{L}_{ce}\left(f_\phi\left(f_b \left(x; p_I^{(1)}, p_s^1\right)\right), y\right),
\label{eq: obj-t1}
\end{equation}
where $\mathcal{L}_{ce}$ denotes the cross-entropy loss. 
Before moving to the next time step, we store all prompts and the associated keys into the prompt bank $\mathcal{P}^1=\{p_I^{(1)}, \left[{p}_s^{1},k_1\right]\}$.
%
The domain-specific prompt and its key value would be isolated from further changes while the domain-invariant prompt would be passed to the next time step to iteratively evolve.

For the subsequent time step $t~(t>1)$, we keep the classification model (including $f_{\phi}$ and $f_{b}$) frozen, and optimize the domain-specific prompt $p_s^t$ and the domain-invariant prompt $p_I^{(t)}$ asynchronously in separate steps.
We first learn an independent domain-specific prompt $p_s^t$ from scratch with the old domain-invariant prompt $p_I^{(t-1)}$ fixed by
\begin{equation} 
\min _{p_s^t}\ \mathcal{L}_{ce}\left(f_{\phi}\left(f_b\left(x; p_I^{(t-1)}, p_s^t\right)\right), y\right),
\label{eq: obj-t-dsp}
\end{equation}
where $(x,y) \in D_t$.
Then we update the domain-invariant prompt by style-augmented prompt refining, which would be thoroughly described in the following subsection.

\subsection{Style-augmented Prompt Refining}
\label{style-augmen}

The straightforward way to update the domain-invariant prompt is to finetune the previous one $p_I^{(t-1)}$ along with the $t$-th domain-specific prompt $p^t_s$.
However, this would easily make the latest domain-invariant prompt not compatible with early domain-specific prompts recorded in the prompt bank.
To address this issue, we build a graph with all existing prompts and feed it into the graph attention network (GAT)~\cite{velivckovic2017graph} to guide the domain-invariant prompt exploring the co-existing and generic features.

\subsubsection{GAT setup}
We flatten all existing prompts into long vectors as $P^t = \{\mathbf{p}^{(t-1)}_I, \mathbf{p}^1_s, ..., \mathbf{p}^t_s\}$, and take them as the nodes of the graph.
The graph attention network consists of one learnable linear transformation $\mathbf{W}\in \mathbb{R}^{L\times L}$, where $L=l\times m$, and a trainable single-layer feed-forward neural network $a$ to acquire the attention coefficients $e$ between two nodes.
Particularly for the node of the domain-invariant prompt that we most concern, the attention coefficients for its $i$-th neighbor $e_{IS}^i$ and itself $e_{II}$ are computed as
\begin{equation}
\begin{aligned}
    e_{IS}^i &= a\left(\mathbf{W}\mathbf{p}_I^{(t-1)}, \mathbf{W}\mathbf{p}_s^{i}\right),\\
    e_{II} &= a\left(\mathbf{W}\mathbf{p}_I^{(t-1)}, \mathbf{W}\mathbf{p}_I^{(t-1)}\right),
\end{aligned}
\end{equation}
which indicates the correlation and relevance between the domain-invariant prompt and each prompt in the bank.
We normalize the above coefficients as $\alpha_{IS}^i$ and $\alpha_{II}$, and use them to adjust the participation of each node in the knowledge aggregation to the domain-invariant prompt as
\begin{equation}
\begin{aligned}
    \alpha_{IS}^i &= \frac{\exp\left(e_{IS}^i\right)}{ \exp\left(e_{II}\right) + \sum_{i=1}^t \exp\left(e_{IS}^i\right)},\\
    \alpha_{II} &= \frac{\exp\left(e_{II}\right)}{ \exp\left(e_{II}\right) + \sum_{i=1}^t \exp\left(e_{IS}^i\right)},
\end{aligned}
\end{equation}
The outputs of the domain-invariant prompt $\mathbf{p}_I^{(t)}$ would be
\begin{equation}
\mathbf{p}_I^{(t)}=f_{GAT}(P^t)=\sum_{i=1}^t\alpha_{IS}^i\mathbf{W}\mathbf{p}_s^{i} + \alpha_{II}\mathbf{W}\mathbf{p}_I^{(t-1)},
\label{eq: GAT-DIP}
\end{equation}
where $f_{GAT}$ denotes the graph attention network. By simply reshaping $\mathbf{p}_I^{(t)}$ into the original prompt size, we could obtain the updated domain-invariant prompt $p_I^{(t)}$.

\subsubsection{Style-augmented GAT training}
\begin{figure}[!t]
    \centering
    \includegraphics[width=0.4\textwidth]{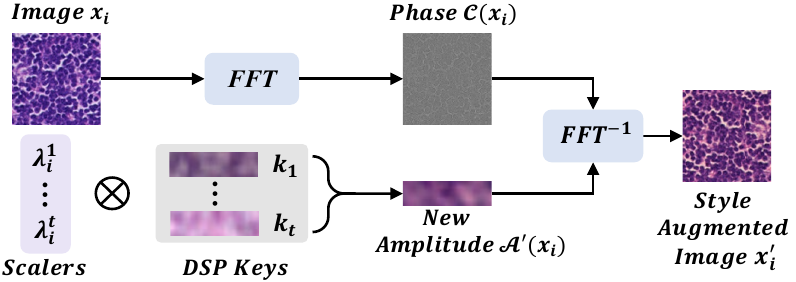}
    \caption{Illustration of generating style-augmented data.}
\label{fig:FFT}
\end{figure}

We augment the style diversity met in GAT training to further improve the generalization potential of the domain-invariant prompt.
As aforementioned, for each image, its amplitude spectrum reflects the low-level features like style or appearance, while the phase spectrum presents its high-level content like shape.
Given an image-label pair of the current domain $(x_i, y_i)\in D_{t}$, we reserve the phrase spectrum $\mathcal{C}(x_i)$ to keep its semantic content, but substitute its amplitude $\mathcal{A}(x_i)$ into a new one $\mathcal{A'}(x_i)$ to modulate its style.
As shown in Fig.~\ref{fig:FFT}, rather than blindly guessing feasible appearances, we use a set of random scalars $\left\{\lambda_i^1, ..., \lambda_i^t\right\}$ to interpolate the average amplitudes of past domains (\ie, the keys in the prompt bank) and generate the new amplitude $\mathcal{A'}(x_i)$ as
\begin{equation}
    \mathcal{A'}(x_i)=\lambda_i^1 k_1+\lambda_i^2  k_2+\cdots+\lambda_i^t k_t, 
\label{eq: aug-new-amp}
\end{equation}
where $\sum_{j=0}^t\lambda_i^j = 1$.
We then perform inverse fast Fourier transform operation $\Phi_{\text{FFT}}^{-1}$ to remap the phase $\mathcal{C}(x_i)$ and amplitude $\mathcal{A'}(x_i)$ into the image space, and generate the style-augmented image $x'_i$ as follows
\begin{equation} 
    x'_i=\Phi_{\text{FFT}}^{-1}\left(\mathcal{A'}(x_i), \mathcal{C}(x_i)\right).
\label{eq: aug-fft-ivers}
\end{equation}

We paired the style-augmented image $x'_i$ with its original label $y_i$ to form a set $D_{t}^{sa} = \left\{(x'_i, y_i), i\in[1, N_t]\right\}$, which will be used for GAT training along with the current domain $D_t$.
For any data $(x,y) \in D_t\cup D_{t}^{sa}$, we select the most compatible domain-specific prompt $p_s^{j*}$ by similarity ranking as 
\begin{equation}
    j^* = {\operatorname{argmax}}_j\  \gamma\left(\mathcal{A}(x), k_j\right),
\label{eq: select-dsp}
\end{equation}
where $j \in [1, t]$ and $\gamma$ denotes the cosine similarity.
Then,  we force the GAT to produce a domain-invariant prompt $p_I^t$ that could satisfyingly tackle the images of any augmented style and work smoothly with any domain-specific prompts in the bank by the following objectives
\begin{equation}
    \min_{f_{GAT}}\ \mathcal{L}_{ce}\left(f_{\phi}\left(f_b\left(x; f_{GAT}(P^{t}),p_s^{j*}\right)\right), y\right).
\label{eq: obj-t-dip}
\end{equation}

The overall training scheme is presented in Algorithm~\ref{algorithm}. 
During inference, we use the latest domain-invariant prompt for all test data and pair each test sample with the most compatible domain-specific prompt to it by Eq.~\ref{eq: select-dsp} accordingly.
%
%
%
%
\section{Experiment}
\subsection{Dataset and Experiment Settings}
\subsubsection{Breast cancer metastase classification}
We adopted the Camelyon17 dataset~\cite{bandi2018detection} which
provided the labels of the presence or absence of breast cancer.
The data was collected from 5 medical centers with different stains.
We closely followed the domain split of prior works~\cite{jiang2022harmofl} and took the samples of the same center as one domain.
All domains are sequentially delivered one by one in ascending order.
We set the total time step as 4, where Domain 4 currently arrives, Domain 1-3 are previously delivered, and Domain 5 remains unseen to the model.

\subsubsection{Epithelium-stroma tissue classification}
We utilized four public datasets, including 615 images from VGH~\cite{beck2011systematic} (Domain 1), 671 images from NKI~\cite{beck2011systematic} (Domain 2), 1296 patches from IHC~\cite{linder2012identification} (Domain 3), and 26,437 patches from NCH~\cite{kather2019predicting} (Domain 4). 
Each of them comes from different institutions under different H\&E stains.
Here, we set the total time step as 3, where Domain 3 has currently arrived, Domain 1 and 2 are previously delivered and Domain 4 remains unseen during model training.

\subsubsection{Implementation details}
We adopted the ViT-B/16~\cite{dosovitskiy2020image} as our feature extractor $f_b$.
We employ the Adam optimizer with the learning rate of $7.5e^{-4}$ in the first time step and the learning rate of $1e^{-4}$ for the subsequent time steps.
%
\subsubsection{Evaluation metrics}
We employed the classification accuracy (Acc) as the base evaluation metric.
We first measure the model performance at the last incremental learning step on all domains, including previous domains, the current domain and unseen domains, to extensively evaluate the ability to alleviate forgetting and generalize.
We further employ three more metrics to 
comprehensively evaluate the overall performance of the entire incremental learning span.
Backward transfer (BWT) evaluates the model stability, \ie, the ability to alleviate catastrophic forgetting, which is computed as
$BWT=\sfrac{{2}\sum_{i=2}^N \sum_{j=1}^{i-1}\left(R_{i, j}-R_{j, j}\right)}{{N(N-1)}}$, where $R_{i, j}$ denotes the Acc of the model trained sequentially from the 1st domain to the $i$-th domain and tested on the $j$-th domain, and $N$ denotes the total number of training domains.
Incremental learning (IL) provides an overall measurement of all seen domains and measures both stability and plasticity of the model as $IL = \sfrac{2\sum_{i=1}^N \sum_{j=1}^{i}R_{i,j}}{N(N+1)}$.
Forward Transfer on unseen domains (FTU) measures the generalization ability of the model as $FTU = \sfrac{2\sum_{i=1}^N \sum_{j=i+1}^{N}R_{i,j}}{N(N-1)}$.

\begin{algorithm}[t]
\caption{Training Procedures} 
\label{algorithm}
{\bf Output:} 
The model $f_{\phi}(f_b (\cdot))$ and prompt bank $\mathcal{P}$.
\\
\While{incrementally learning from $t=1$ to $T$}{
    \eIf{$t == 1$}{
        Load pre-trained weights in $f_b$ and freeze it.\\
        Optimize $p_I^{(1)}$, $p_s^1$, $f_{\phi}$ with $D_1$ by Eq.~\ref{eq: obj-t1}.\\
        Calculate the key $k_1$ for ${p}_s^{1}$ by Eq.~\ref{eq: dsp-t-key}. \\ 
        Store all prompts in the bank $\mathcal{P}^1$.\\ 
    }{
        Freeze $f_b$, $f_{\phi}$ and train $p_s^t$ with $D_t$ by Eq.~\ref{eq: obj-t-dsp}.\\
        Generate style-augmented data $D_t^{sa}$ by Eq.~\ref{eq: aug-new-amp}.\\
        Update $p_I^{(t)}$ given $p_s^t$ by GAT as  Eq.~\ref{eq: obj-t-dip}.\\
        Compute the key $k_t$ by Eq.~\ref{eq: aug-new-amp} and append $[k_t, p_s^t]$ in the prompt bank $\mathcal{P}^t$.\\
        Overwrite DIP as $p_I^{(t)}$ in the prompt bank $\mathcal{P}^t$.\\ 
    }
    Pass $f_b$, $f_{\phi}$ and $\mathcal{P}^t$ in the $t+1$ step.
}
{\bf Return} $f_{\phi}(f_b(\cdot))$, $\mathcal{P} \leftarrow \mathcal{P}^T$.
\end{algorithm}

\subsection{Comparison with the State-of-the-arts}

\begin{table*}[t]
\fontsize{9pt}\baselineskip\selectfont
\centering
\begin{tabular}{l|c|c|c|c|c|c|c|c|c}
   \toprule[1pt]
   \multirow{2}{*}{ Methods } & \multicolumn{6}{c|}{ Acc [\%] $\uparrow$} & \multirow{2}{*}{IL $\uparrow$ } & \multirow{2}{*}{BWT $\uparrow$} & \multirow{2}{*}{FTU $\uparrow$}
\\\cline{2-7}
     & Domain 1 & Domain 2 & Domain 3 & Domain 4 & Domain 5 & Avg & & & 
\\
   \midrule
   Individual Training & \begin{tabular}[c]{@{}l@{}}~96.79 \\($\pm$0.77) \end{tabular} & \begin{tabular}[c]{@{}l@{}}~91.62 \\($\pm$0.82) \end{tabular} & \begin{tabular}[c]{@{}l@{}}~96.32 \\($\pm$0.54) \end{tabular} & \begin{tabular}[c]{@{}l@{}}~96.46 \\($\pm$0.65) \end{tabular} & \begin{tabular}[c]{@{}l@{}}~96.84 \\($\pm$0.70) \end{tabular} & \begin{tabular}[c]{@{}l@{}}~95.30 \\($\pm$0.52) \end{tabular} & \begin{tabular}[c]{@{}l@{}}~71.74 \\($\pm$0.56) \end{tabular} & \begin{tabular}[c]{@{}l@{}}~-38.65 \\($\pm$0.55) \end{tabular}  & \begin{tabular}[c]{@{}l@{}}~64.71 \\($\pm$1.31) \end{tabular}
\\ 
   \begin{tabular}[c]{@{}l@{}}Joint Training\\ (Upper bound)\end{tabular}& \begin{tabular}[c]{@{}l@{}}~96.96 \\($\pm$0.54) \end{tabular} & \begin{tabular}[c]{@{}l@{}}~94.32 \\($\pm$0.67) \end{tabular} & \begin{tabular}[c]{@{}l@{}}~97.51 \\($\pm$0.52) \end{tabular} & \begin{tabular}[c]{@{}l@{}}~97.41 \\($\pm$0.63) \end{tabular} & \begin{tabular}[c]{@{}l@{}}~82.80 \\($\pm$1.19) \end{tabular} & \begin{tabular}[c]{@{}l@{}}~92.45 \\($\pm$0.58) \end{tabular} & \begin{tabular}[c]{@{}l@{}}~95.27 \\($\pm$0.46) \end{tabular} & \begin{tabular}[c]{@{}l@{}}~2.53 \\($\pm$0.03) \end{tabular} & \begin{tabular}[c]{@{}l@{}}~80.75 \\($\pm$1.22) \end{tabular}
\\
   \midrule
   Sequential Finetune & \begin{tabular}[c]{@{}l@{}}~61.92 \\($\pm$1.22) \end{tabular} & \begin{tabular}[c]{@{}l@{}}~64.78 \\($\pm$2.55) \end{tabular} & \begin{tabular}[c]{@{}l@{}}~51.52 \\($\pm$1.10) \end{tabular} & \begin{tabular}[c]{@{}l@{}}~96.51 \\($\pm$0.68) \end{tabular} & \begin{tabular}[c]{@{}l@{}}~49.82 \\($\pm$3.22) \end{tabular} & \begin{tabular}[c]{@{}l@{}}~64.91 \\($\pm$2.84) \end{tabular} & \begin{tabular}[c]{@{}l@{}}~72.59 \\($\pm$2.01) \end{tabular} & \begin{tabular}[c]{@{}l@{}}~-37.13 \\($\pm$1.69) \end{tabular} & \begin{tabular}[c]{@{}l@{}}~63.69 \\($\pm$2.62) \end{tabular}
\\
   LwF & \begin{tabular}[c]{@{}l@{}}~83.28 \\($\pm$0.44) \end{tabular} & \begin{tabular}[c]{@{}l@{}}~72.69 \\($\pm$1.05) \end{tabular} & \begin{tabular}[c]{@{}l@{}}~59.71 \\($\pm$0.65) \end{tabular} & \begin{tabular}[c]{@{}l@{}}~95.86 \\($\pm$0.83) \end{tabular} & \begin{tabular}[c]{@{}l@{}}~69.07 \\($\pm$0.53) \end{tabular} & \begin{tabular}[c]{@{}l@{}}~76.12 \\($\pm$0.68) \end{tabular} & \begin{tabular}[c]{@{}l@{}}~80.12 \\($\pm$1.11) \end{tabular} & \begin{tabular}[c]{@{}l@{}}~-23.44 \\($\pm$0.36) \end{tabular}  & \begin{tabular}[c]{@{}l@{}}~69.18 \\($\pm$0.47) \end{tabular}
\\
   EWC & \begin{tabular}[c]{@{}l@{}}~81.87 \\($\pm$0.81) \end{tabular} & \begin{tabular}[c]{@{}l@{}}~77.21 \\($\pm$1.44) \end{tabular} & \begin{tabular}[c]{@{}l@{}}~53.26 \\($\pm$2.21) \end{tabular} & \begin{tabular}[c]{@{}l@{}}~96.47 \\($\pm$0.99) \end{tabular} & \begin{tabular}[c]{@{}l@{}}~46.77 \\($\pm$2.55) \end{tabular} & \begin{tabular}[c]{@{}l@{}}~71.12 \\($\pm$2.28) \end{tabular} & \begin{tabular}[c]{@{}l@{}}~77.78 \\($\pm$2.46) \end{tabular} & \begin{tabular}[c]{@{}l@{}}~-29.32 \\($\pm$0.48) \end{tabular}  & \begin{tabular}[c]{@{}l@{}}~61.02 \\($\pm$2.26) \end{tabular}
\\
   SI & \begin{tabular}[c]{@{}l@{}}~86.53 \\($\pm$1.66) \end{tabular} & \begin{tabular}[c]{@{}l@{}}~82.49 \\($\pm$0.97) \end{tabular} & \begin{tabular}[c]{@{}l@{}}~54.04 \\($\pm$1.78) \end{tabular} & \begin{tabular}[c]{@{}l@{}}~96.52 \\($\pm$0.64) \end{tabular} & \begin{tabular}[c]{@{}l@{}}~50.55 \\($\pm$1.43) \end{tabular} & \begin{tabular}[c]{@{}l@{}}~74.03 \\($\pm$1.21) \end{tabular} & \begin{tabular}[c]{@{}l@{}}~79.83 \\($\pm$1.29) \end{tabular} & \begin{tabular}[c]{@{}l@{}}~-24.15 \\($\pm$0.61) \end{tabular}  & \begin{tabular}[c]{@{}l@{}}~62.51 \\($\pm$1.05) \end{tabular}
\\
   DGR & \begin{tabular}[c]{@{}l@{}}~81.21 \\($\pm$2.49) \end{tabular} & \begin{tabular}[c]{@{}l@{}}~83.20 \\($\pm$1.96) \end{tabular} & \begin{tabular}[c]{@{}l@{}}~72.77 \\($\pm$3.45) \end{tabular} & \cellcolor{lightgray}\begin{tabular}[c]{@{}l@{}}~\textbf{96.84} \\($\pm$1.10) \end{tabular} & \begin{tabular}[c]{@{}l@{}}~78.57 \\($\pm$2.83) \end{tabular}& \begin{tabular}[c]{@{}l@{}}~82.52 \\($\pm$2.36) \end{tabular} & \begin{tabular}[c]{@{}l@{}}~86.80 \\($\pm$1.95) \end{tabular} &\begin{tabular}[c]{@{}l@{}}~-15.67 \\($\pm$1.62) \end{tabular}  & \begin{tabular}[c]{@{}l@{}}~70.64 \\($\pm$2.18) \end{tabular}
\\
   Orc-MML & \begin{tabular}[c]{@{}l@{}}~90.07 \\($\pm$0.93) \end{tabular} & \begin{tabular}[c]{@{}l@{}}~86.64 \\($\pm$0.85) \end{tabular} & \begin{tabular}[c]{@{}l@{}}~87.32 \\($\pm$0.77) \end{tabular} & \begin{tabular}[c]{@{}l@{}}~89.73 \\($\pm$0.96) \end{tabular} & \begin{tabular}[c]{@{}l@{}}~66.91 \\($\pm$1.59) \end{tabular} & \begin{tabular}[c]{@{}l@{}}~84.13 \\($\pm$0.91) \end{tabular} & \begin{tabular}[c]{@{}l@{}}~89.69 \\($\pm$0.94) \end{tabular} & \begin{tabular}[c]{@{}l@{}}~-6.44 \\($\pm$0.60) \end{tabular} & \begin{tabular}[c]{@{}l@{}}~75.45 \\($\pm$1.03) \end{tabular} 
\\
   S-Prompt & \begin{tabular}[c]{@{}l@{}}~92.50 \\($\pm$0.69) \end{tabular} & \begin{tabular}[c]{@{}l@{}}~81.30 \\($\pm$0.88) \end{tabular} & \begin{tabular}[c]{@{}l@{}}~93.93 \\($\pm$0.70) \end{tabular} & \begin{tabular}[c]{@{}l@{}}~94.43 \\($\pm$0.83) \end{tabular} & \begin{tabular}[c]{@{}l@{}}~74.37 \\($\pm$0.92) \end{tabular} & \begin{tabular}[c]{@{}l@{}}~87.31 \\($\pm$0.85) \end{tabular} & \begin{tabular}[c]{@{}l@{}}~91.73 \\($\pm$0.78) \end{tabular} & \begin{tabular}[c]{@{}l@{}}~-3.46 \\($\pm$0.24) \end{tabular} & \begin{tabular}[c]{@{}l@{}}~80.64 \\($\pm$0.62) \end{tabular}
\\
   DualPrompt & \begin{tabular}[c]{@{}l@{}}~91.34 \\($\pm$0.43) \end{tabular} & \begin{tabular}[c]{@{}l@{}}~85.18 \\($\pm$0.75) \end{tabular} & \begin{tabular}[c]{@{}l@{}}~92.71 \\($\pm$0.67) \end{tabular} & \begin{tabular}[c]{@{}l@{}}~95.63 \\($\pm$0.58) \end{tabular} & \begin{tabular}[c]{@{}l@{}}~67.44 \\($\pm$2.78) \end{tabular} & \begin{tabular}[c]{@{}l@{}}~86.46 \\($\pm$0.59) \end{tabular} & \begin{tabular}[c]{@{}l@{}}~91.11 \\($\pm$0.42) \end{tabular} & \begin{tabular}[c]{@{}l@{}}~-7.60 \\($\pm$0.11) \end{tabular}  & \begin{tabular}[c]{@{}l@{}}~77.57 \\($\pm$1.02) \end{tabular}
\\
   \midrule
   Ours & \cellcolor{lightgray}\begin{tabular}[c]{@{}l@{}}~\textbf{94.36} \\ ($\pm$0.90)\end{tabular} & \cellcolor{lightgray}\begin{tabular}[c]{@{}l@{}}~\textbf{89.40} \\($\pm$0.39) \end{tabular} &\cellcolor{lightgray}\begin{tabular}[c]{@{}l@{}}~\textbf{94.32} \\($\pm$0.87) \end{tabular} & \begin{tabular}[c]{@{}l@{}}~95.86 \\($\pm$0.79) \end{tabular} & \cellcolor{lightgray}\begin{tabular}[c]{@{}l@{}}~\textbf{84.12} \\($\pm$1.26) \end{tabular} & \cellcolor{lightgray}\begin{tabular}[c]{@{}l@{}}~\textbf{91.61} \\($\pm$0.67) \end{tabular} & \cellcolor{lightgray}\begin{tabular}[c]{@{}l@{}}~\textbf{93.67} \\($\pm$0.44) \end{tabular} & \cellcolor{lightgray}\begin{tabular}[c]{@{}l@{}}~\textbf{-1.62} \\($\pm$0.09) \end{tabular} & \cellcolor{lightgray}\begin{tabular}[c]{@{}l@{}}~\textbf{82.17} \\($\pm$0.54) \end{tabular}
\\
   \bottomrule[1pt]
\end{tabular}
\caption{
The comparison results of breast cancer metastases classification in the final time step (the 2-7 Columns) and the entire domain incremental learning process (the last three columns). We have highlighted the best DIL results in bold.
}
\label{table: 1-N-T-clsf}
\end{table*}

\begin{table*}[t]
\fontsize{10pt}\baselineskip\selectfont
\centering
\begin{tabular}{l|c|c|c|c|c|c|c|c}
   \toprule[1pt]
   \multirow{2}{*}{ Methods } & \multicolumn{5}{c|}{ Acc [\%] $\uparrow$}  & \multirow{2}{*}{IL $\uparrow$} & \multirow{2}{*}{BWT $\uparrow$} & \multirow{2}{*}{FTU $\uparrow$}
\\\cline{2-6}
     & Domain 1 & Domain 2 & Domain 3 & Domain 4 & Avg & & 
\\
   \midrule
   Individual Training & \begin{tabular}[c]{@{}l@{}}~94.09 \\($\pm$0.56 \end{tabular} & \begin{tabular}[c]{@{}l@{}}~91.84 \\($\pm$0.61 \end{tabular} & \begin{tabular}[c]{@{}l@{}}~93.42 \\($\pm$0.83) \end{tabular} & \begin{tabular}[c]{@{}l@{}}~96.50 \\($\pm$0.69) \end{tabular} & \begin{tabular}[c]{@{}l@{}}~93.96 \\($\pm$0.60) \end{tabular} & \begin{tabular}[c]{@{}l@{}}~78.79 \\($\pm$0.64) \end{tabular} & \begin{tabular}[c]{@{}l@{}}~-28.87 \\($\pm$0.35) \end{tabular} & \begin{tabular}[c]{@{}l@{}}~67.74 \\($\pm$2.19) \end{tabular} 
\\
   \begin{tabular}[c]{@{}l@{}}Joint Training\\ (Upper bound)\end{tabular} &  \begin{tabular}[c]{@{}l@{}}~94.33 \\($\pm$0.48) \end{tabular} & \begin{tabular}[c]{@{}l@{}}~92.61 \\($\pm$0.65) \end{tabular} & \begin{tabular}[c]{@{}l@{}}~93.67 \\($\pm$0.71) \end{tabular} & \begin{tabular}[c]{@{}l@{}}~85.82 \\($\pm$1.86) \end{tabular} & \begin{tabular}[c]{@{}l@{}}~91.61 \\($\pm$0.84) \end{tabular} & \begin{tabular}[c]{@{}l@{}}~93.56 \\($\pm$0.52) \end{tabular} & \begin{tabular}[c]{@{}l@{}}~1.20 \\($\pm$0.04) \end{tabular} & \begin{tabular}[c]{@{}l@{}}~74.91 \\($\pm$0.93) \end{tabular} \\
   \midrule
   Sequential Finetune & \begin{tabular}[c]{@{}l@{}}~62.77 \\($\pm$1.19) \end{tabular} & \begin{tabular}[c]{@{}l@{}}~70.14 \\($\pm$2.06) \end{tabular} & \begin{tabular}[c]{@{}l@{}}~93.58 \\($\pm$0.95) \end{tabular} & \begin{tabular}[c]{@{}l@{}}~69.06 \\($\pm$1.42) \end{tabular} & \begin{tabular}[c]{@{}l@{}}~73.89 \\($\pm$1.27) \end{tabular} & \begin{tabular}[c]{@{}l@{}}~82.84 \\($\pm$1.44) \end{tabular} & \begin{tabular}[c]{@{}l@{}}~-20.86 \\($\pm$1.08) \end{tabular} & \begin{tabular}[c]{@{}l@{}}~66.02 \\($\pm$1.79) \end{tabular}
\\
   LwF & \begin{tabular}[c]{@{}l@{}}~77.54 \\($\pm$1.60) \end{tabular} & \begin{tabular}[c]{@{}l@{}}~66.32 \\($\pm$2.47) \end{tabular} & \begin{tabular}[c]{@{}l@{}}~93.52 \\($\pm$1.01) \end{tabular} & \begin{tabular}[c]{@{}l@{}}~73.61 \\($\pm$1.99) \end{tabular} & \begin{tabular}[c]{@{}l@{}}~77.75 \\($\pm$1.32) \end{tabular} & \begin{tabular}[c]{@{}l@{}}~85.12 \\($\pm$1.17) \end{tabular} & \begin{tabular}[c]{@{}l@{}}~-15.82 \\($\pm$0.91) \end{tabular} & \begin{tabular}[c]{@{}l@{}}~73.93 \\($\pm$1.85) \end{tabular}
\\
   EWC & \begin{tabular}[c]{@{}l@{}}~71.22 \\($\pm$2.04) \end{tabular} & \begin{tabular}[c]{@{}l@{}}~77.90 \\($\pm$1.70) \end{tabular} & \begin{tabular}[c]{@{}l@{}}~93.61 \\($\pm$0.84) \end{tabular} & \begin{tabular}[c]{@{}l@{}}~68.12 \\($\pm$2.66) \end{tabular} & \begin{tabular}[c]{@{}l@{}}~77.71 \\($\pm$1.82) \end{tabular} & \begin{tabular}[c]{@{}l@{}}~87.85 \\($\pm$1.79) \end{tabular} & \begin{tabular}[c]{@{}l@{}}~-15.01 \\($\pm$0.38) \end{tabular} & \begin{tabular}[c]{@{}l@{}}~72.32 \\($\pm$1.88) \end{tabular}
\\
   SI & \begin{tabular}[c]{@{}l@{}}~66.58 \\($\pm$0.47) \end{tabular} & \begin{tabular}[c]{@{}l@{}}~72.51 \\($\pm$0.21) \end{tabular} & \begin{tabular}[c]{@{}l@{}}~93.55 \\($\pm$1.02) \end{tabular} & \begin{tabular}[c]{@{}l@{}}~77.33 \\($\pm$1.55) \end{tabular} & \begin{tabular}[c]{@{}l@{}}~77.50 \\($\pm$0.68) \end{tabular} & \begin{tabular}[c]{@{}l@{}}~82.64 \\($\pm$0.64) \end{tabular} & \begin{tabular}[c]{@{}l@{}}~-21.99 \\($\pm$0.82) \end{tabular} & \begin{tabular}[c]{@{}l@{}}~71.51 \\($\pm$0.73) \end{tabular}
\\
   DGR & \begin{tabular}[c]{@{}l@{}}~80.54 \\($\pm$1.30) \end{tabular} & \begin{tabular}[c]{@{}l@{}}~82.35 \\($\pm$1.77) \end{tabular} & \begin{tabular}[c]{@{}l@{}}~93.67 \\($\pm$0.69) \end{tabular} & \begin{tabular}[c]{@{}l@{}}~86.72 \\($\pm$1.48) \end{tabular} & \begin{tabular}[c]{@{}l@{}}~85.81 \\($\pm$0.74) \end{tabular} & \begin{tabular}[c]{@{}l@{}}~88.80 \\($\pm$0.60) \end{tabular} & \begin{tabular}[c]{@{}l@{}}~-9.72 \\($\pm$0.18) \end{tabular} & \begin{tabular}[c]{@{}l@{}}~78.29 \\($\pm$1.95) \end{tabular}
\\
   Orc-MML & \begin{tabular}[c]{@{}l@{}}~85.61 \\($\pm$0.93) \end{tabular} & \begin{tabular}[c]{@{}l@{}}~84.93 \\($\pm$0.75) \end{tabular} & \begin{tabular}[c]{@{}l@{}}~83.76 \\($\pm$0.67) \end{tabular} & \begin{tabular}[c]{@{}l@{}}~72.58 \\($\pm$0.99) \end{tabular} & \begin{tabular}[c]{@{}l@{}}~81.72 \\($\pm$0.72) \end{tabular} & \begin{tabular}[c]{@{}l@{}}~88.15 \\($\pm$0.71) \end{tabular} & \begin{tabular}[c]{@{}l@{}}~-6.82 \\($\pm$0.05) \end{tabular} & \begin{tabular}[c]{@{}l@{}}~69.19 \\($\pm$1.26) \end{tabular} 
\\
   S-Prompt & \begin{tabular}[c]{@{}l@{}}~85.29 \\($\pm$0.83) \end{tabular} & \begin{tabular}[c]{@{}l@{}}~88.42 \\($\pm$1.94) \end{tabular} & \cellcolor{lightgray}\begin{tabular}[c]{@{}l@{}}~\textbf{94.33} \\($\pm$0.75) \end{tabular} & \begin{tabular}[c]{@{}l@{}}~73.03 \\($\pm$1.76) \end{tabular} & \begin{tabular}[c]{@{}l@{}}~85.27 \\($\pm$0.94) \end{tabular} & \begin{tabular}[c]{@{}l@{}}~89.61 \\($\pm$1.04) \end{tabular} & \begin{tabular}[c]{@{}l@{}}~-4.19 \\($\pm$0.06) \end{tabular} & \begin{tabular}[c]{@{}l@{}}~74.62 \\($\pm$1.01) \end{tabular} 
\\
   DualPrompt & \begin{tabular}[c]{@{}l@{}}~88.94 \\($\pm$0.58) \end{tabular} & \begin{tabular}[c]{@{}l@{}}~88.05 \\($\pm$0.47) \end{tabular} & \begin{tabular}[c]{@{}l@{}}~93.63 \\($\pm$0.17) \end{tabular} & \begin{tabular}[c]{@{}l@{}}~83.16 \\($\pm$0.51) \end{tabular} & \begin{tabular}[c]{@{}l@{}}~88.45 \\($\pm$0.33) \end{tabular} & \begin{tabular}[c]{@{}l@{}}~91.14 \\($\pm$0.77) \end{tabular} & \begin{tabular}[c]{@{}l@{}}~-4.34 \\($\pm$0.03) \end{tabular} & \begin{tabular}[c]{@{}l@{}}~75.11 \\($\pm$0.42) \end{tabular}
\\ \midrule
   Ours & \cellcolor{lightgray}\begin{tabular}[c]{@{}l@{}}~\textbf{90.61} \\($\pm$1.00) \end{tabular} & \cellcolor{lightgray}\begin{tabular}[c]{@{}l@{}}~\textbf{88.47} \\($\pm$0.39) \end{tabular} & \begin{tabular}[c]{@{}l@{}}~93.84 \\($\pm$1.03) \end{tabular} & \cellcolor{lightgray}\begin{tabular}[c]{@{}l@{}}~\textbf{89.14} \\($\pm$0.97) \end{tabular} & \cellcolor{lightgray}\begin{tabular}[c]{@{}l@{}}~\textbf{90.52} \\($\pm$0.55) \end{tabular} & \cellcolor{lightgray}\begin{tabular}[c]{@{}l@{}}~\textbf{92.17} \\($\pm$0.49) \end{tabular} & \cellcolor{lightgray}\begin{tabular}[c]{@{}l@{}}~\textbf{-2.19} \\($\pm$0.05) \end{tabular} & \cellcolor{lightgray}\begin{tabular}[c]{@{}l@{}}~\textbf{80.06} \\($\pm$0.84) \end{tabular}
\\
   \bottomrule[1pt]
\end{tabular}
\caption{
The comparison results of epithelium-stroma classification in the final time step (the 2-6 columns) and the entire domain incremental learning process (the last three columns). We have highlighted the best DIL results in bold.
}
\label{table: 2-E-S-clsf}
\end{table*}

\subsubsection{Compared methods}
Besides the intuitive method, \ie, sequential finetune, we also implemented the state-of-the-art incremental learning methods, including the regularization-based methods (LwF~\cite{li2017learning}, EWC~\cite{kirkpatrick2017overcoming} and SI~\cite{zenke2017continual}), the replay-based method (DGR~\cite{shin2017continual}), and the parameter isolation methods (Orc-MML~\cite{gonzalez2020wrong}) especially those also involved prompt tuning (DualPrompt~\cite{wang2022dualprompt} and S-Prompt~\cite{wang2022s}).
We separately trained a model for each domain (\ie, individual training) and also jointly trained a model with all delivered domains.
Since the joint training could fully access all seen domains while the DIL methods only have access to the current domain, we consider it as the upper bound.

\subsubsection{Experiment results}
We reported the performance of breast cancer metastase classification in Table~\ref{table: 1-N-T-clsf}. Each experiment is repeated 5 times to avoid random bias.
Our approach achieved the best results in the majority of evaluation metrics (8 out of 9). 
Compared to prior parameter isolation approaches like Orc-MML, DualPrompt, and S-Prompt, our framework yielded the least forgetting of past domains with 1.86\% increases in Domain 1 and 2.76\% increases in Domain 2.
Since we continually evolve the domain-invariant prompt by style-augmented prompt refining, our approach greatly improved model generalization capability, leading to 5.55\% gains in the unseen domain (Domain 5) and 1.53\% increases in FTU compared to the state-of-the-art approach.

We presented the results of epithelium-stroma tissue classification in Table \ref{table: 2-E-S-clsf}. 
Our framework outperformed the competing approaches on most of the evaluation metrics (7 out of 8).
For the first delivered domain (Domain 1), which commonly suffered most from catastrophic forgetting, our framework achieved 5.32\% higher than S-Prompt and 1.67\% higher than DualPrompt, demonstrating the effectiveness of our decoupled prompt tuning.
When it comes to the generalization ability, our approach yielded 2.42\% increases in the unseen domain (Domain 4) and 1.77\% increases in FTU over the competing methods.

\subsection{Analysis of the Key Components}
\subsubsection{The tradeoff between model performance and memory efficiency}
We evaluated the memory efficiency by the metric of model size efficiency (MS)~\cite{diaz2018don}, which measures the additional storage used at the time step $t$ compared to the usage in the first time step by computing $MS=\min \left(1, \frac{1}{N}\sum_{i=1}^N \frac{\theta_1}{\theta_i}\right)$, where $\theta_1$ and $\theta_i$ denote the allocated memory spaces to store all necessary modules for the next round of incremental learning in the 1st and $i$-th time step respectively.
We also calculated the absolute value of the average additional memory storage (AAMS) over time as $AAMS = \frac{1}{N}\sum_{i=1}^N |\theta_i - \theta_1|$.

As presented in Table~\ref{table: abl-memory-eff}, the sequential finetune (Seq-FT) method and regularization-based approaches barely require any additional module to support the learning in the next time step.
However, these methods still have large improvement spaces in the aspect of alleviating model forgetting (see Acc and BWT) and enhancing model generalization (see FTU).
The replay-based approaches and parameter isolation approaches, such as Orc-MML (Orc-M), normally consumed extra memory spaces to trade for model performance.
Among them, the prompt-based approaches, \ie, S-Prompt (S-P), DualPrompt (Dual-P), and ours, are the top 3 memory-efficient ones.
With limited additional memory spaces (around 0.57 MB), our approach could bring significant performance gains with 4.30\% in the average Acc, 1.84\% in BWT, and 1.53\% in FTU over prior prompt-based approaches, suggesting it is the most desirable approach when considering the trade-off between model accuracy and memory consumption.

For training efficiency, our work used 0.6h longer training time than other prompt-based methods on average, which is generally affordable in most cases.
\subsubsection{The efficiency of decoupled prompt tuning}

\begin{table}[t]
\fontsize{9pt}\baselineskip\selectfont
\centering
\begin{tabular}{l|c|c|c|c|c}
   \toprule[1pt]
    Methods & A-Acc $\uparrow$ & BWT $\uparrow$ & FTU $\uparrow$ & AAMS $\downarrow$ & MS $\uparrow$ 
\\
 \midrule
   Seq-FT  & 64.91 & -37.13 & 63.69 & 0 & 1
\\
   LwF & 76.12 & -23.44 & 69.18 & 0 & 1 
\\
   EWC & 71.12 & -29.32 & 61.02 & 0 & 1
\\
   SI & 74.03 & -24.15 & 62.51 & 0 & 1 
\\
   DGR & 82.52 & -15.67 & 70.64 & 399.61 & 0.56
\\
   Orc-M & 84.13 & -6.44 & 75.45 & 460.50 & 0.52 
\\
   S-P & 87.31 & -3.46 & 80.64 & 0.23 & 0.99
\\
   Dual-P & 86.46 & -7.60 & 77.57 & 0.57 & 0.99
\\
   \midrule
   Ours & 91.61 & -1.62 & 82.17 & 0.57 & 0.99
\\
   \bottomrule[1pt]
\end{tabular}
\caption{Analysis of model accuracy and memory efficiency. Here, we reported the average Acc of all domains (A-Acc) in the last time step, and employed model size efficiency (MS) and the average additional memory storage (AAMS) [MB] to measure memory efficiency.}
\label{table: abl-memory-eff}
\end{table}
\begin{table}[t]
\fontsize{9pt}\baselineskip\selectfont
\centering
    \begin{tabular}{l|c|c|c|c|c|c}
    \toprule[1pt]
        DIP & DSP & GAT & SA & A-Acc $\uparrow$ & BWT $\uparrow$   & FTU $\uparrow$   \\ \midrule
        \checkmark &     &     &            & 85.94 & -8.86 & 77.35 \\
         & \checkmark    &     &            & 86.31 & -7.29 & 75.82 \\
        \checkmark & \checkmark    &     &            & 86.24 & -7.16 & 77.49 \\
        \checkmark & \checkmark   & \checkmark   &            & 90.34 & -3.19 & 80.04 \\
        \checkmark & \checkmark & \checkmark   & \checkmark          & 91.61 & -1.62 & 82.17 \\ \bottomrule[1pt]
    \end{tabular}
    \caption{Analysis of the key operations in style-augmented prompt refining.}
    \label{table: abl-dip}
\end{table}

\begin{figure}[t]
    \centering
    \includegraphics[width=0.43\textwidth]{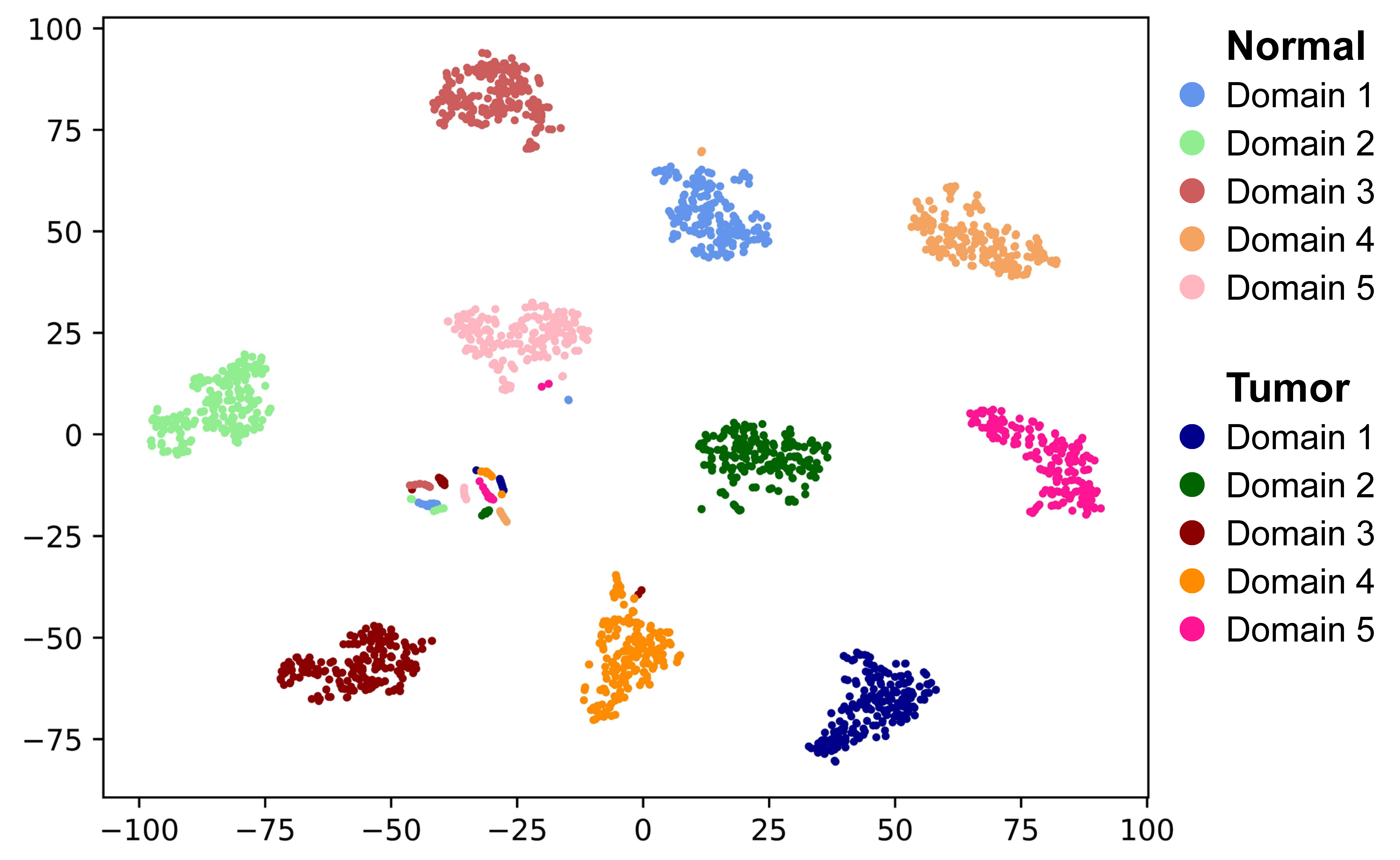}
    \caption{The t-SNE visualization of the feature embeddings after applying decoupled prompt tuning.}
  \label{fig:t-sne}
\end{figure}

We visualized the output feature embeddings after performing decoupled prompt tuning in the breast cancer classification task via t-SNE in Fig.~\ref{fig:t-sne}.
For the embeddings of the same category (\eg, Tumor or Normal), the features within the same domain are grouped into a cluster and well-separated from other domains, suggesting that the learned domain-specific prompts could effectively capture the domain-distinctive characteristics.
For the embeddings within the same domain, it shows a clear decision boundary between
the features of normal tissues and tumor tissues, indicating our model could well distinguish them from each other.

\subsubsection{The study of key operations in style-augmented prompt refining}

To extensively investigate the effectiveness of style-augmented prompt refining, we experimented on several settings, including (a) using DIP individually (first row), (b) using DSP individually (second row), (c) using both DIP and DSP but updating DIP via fine-tuning (third row), and (d) using both DIP and DSP and refining DIP via GAT with the current domain $D_t$ only (fourth row), and compared them with ours, \ie, using both DIP and DSP and refining DIP via GAT with style-augmented training data $D_t^{sa} \cup D_t$ (the last row).
We reported the results on breast cancer classification in Table~\ref{table: abl-dip}.
Compared to simply finetuning DIP, refining with GAT 
could explore more high-correlative and domain-generic representations across domains, leading to 4.10\% increases in the average Acc and 2.55\% increases in FTU.
Further refining by the style-augmented data 
not only keeps the updated DIP compatible with early-recorded DSPs but also lets the model be early-prepared for the unseen styles during inference, thus bringing further improvements of 1.27\% and 2.13\% increases in the average Acc and FTU respectively.
%
%
%
\section{Conclusion}
We presented a memory-efficient prompt tuning framework to incrementally evolve the histology classification model towards a more generalizable and robust direction.
For each incoming domain, we performed decoupled prompt tuning upon the initial classification model with two lightweight prompts, efficiently acquiring the latest domain knowledge without huge memory costs.
We customized a domain-specific prompt customized for tackling the distinctive characteristics while maintaining a domain-invariant prompt shared across all domains to progressively explore the common content embedding.
We additionally conducted style-augmented prompt refining on the domain-invariant prompt to continually investigate domain-generic representations across domains and cultivate its generalization potential.
All prompts will be stored in a prompt bank, where the domain-specific prompts will be isolated from further changes to prevent catastrophic forgetting of past domains, while the domain-invariant prompt will be passed on to the next time step to continually evolve.
We have extensively evaluated our framework with two histology classification tasks, where our approach outperformed other comparison methods with higher accuracy and more satisfying memory efficiency.
%
%
%
\section*{Acknowledgements}
The work described in this paper was supported in part by the following grant from the Research Grants Council of the Hong Kong SAR, China (Project No. T45-401/22-N), the Hong Kong Innovation and Technology Fund (Project No. MHP/085/21), and the National Natural Science Fund (62201483).

\bibliography{0-ref}

\newpage
\setcounter{section}{1}
\section*{Appendix}
\subsection{Dataset and Pre-processing}
\subsubsection{Breast cancer metastase classification}
The data samples of Camelyon17 ~\cite{bandi2018detection} are collected from 5 medical centers, including Radboud University Medical Center (Domain 1), Canisius-Wilhelmina Hospital (Domain 2), University Medical Center Utrecht (Domain 3), Rijnstate Hospital (Domain 4) and Laboratorium Pathologie Oost-Nederland (Domain 5)~\cite{bandi2018detection}.
Each institute contributes 100 annotated whole-slide images (WSIs), which would be cut into 96 $\times$ 96 patches following the preprocessing procedures of Jiang ~\etal~\cite{jiang2022harmofl}.
The data in each domain will be split under the ratio of 6:2:2 for training, validation and testing, respectively.

\subsubsection{Epithelium-stroma tissue classification}
We combined the data of four public datasets, including 615 images from VGH~\cite{beck2011systematic} (Domain 1), 671 images from NKI~\cite{beck2011systematic} (Domain 2), 1296 patches from IHC~\cite{linder2012identification} (Domain 3), and 26,437 patches from NCH~\cite{kather2019predicting} (Domain 4) for the epithelium-stroma tissue classification task.  %
For the VGH and NKI datasets, we closely followed the preprocessing steps in Li~\etal~\cite{lichen2022domain} and sampled the patches with the size of 224$\times$224 from raw images.
Since each dataset is sampled from different institutions under different H\&E stains, we take each dataset as one domain for data heterogeneity.
We split each domain data with the ratio of 6:2:2, and consider them as the training set, validating set, and testing set, respectively.

\subsection{Implementation Details}
We implemented the basic transformer feature extractor $f_b$ closely following the ViT-B/16~\cite{dosovitskiy2020image}.
In the first time step, we used the Adam optimizer with a learning rate of $7.5e^{-4}$ and momentum of 0.9 to train DIP, DSP and classification layer $f_{\phi}$ simultaneously.
For the subsequent time steps, we optimized the DSP and DIP in separate steps.
We first trained the DSP from scratch with the Adam optimizer under the same learning rate and momentum as before.
Then we optimized the DIP upon the previous one by training the GAT with another Adam optimizer in the learning rate of $1e^{-4}$ and momentum of 0.9.
%
%
All experiments are implemented in Pytorch using one Nvidia GTX3090 GPU.

\subsection{Hyper-parameter Analysis}
\subsubsection{Analysis for the optimal positions to attach prompts}

\begin{figure}[th]
    \centering
    \includegraphics[width=0.45\textwidth]{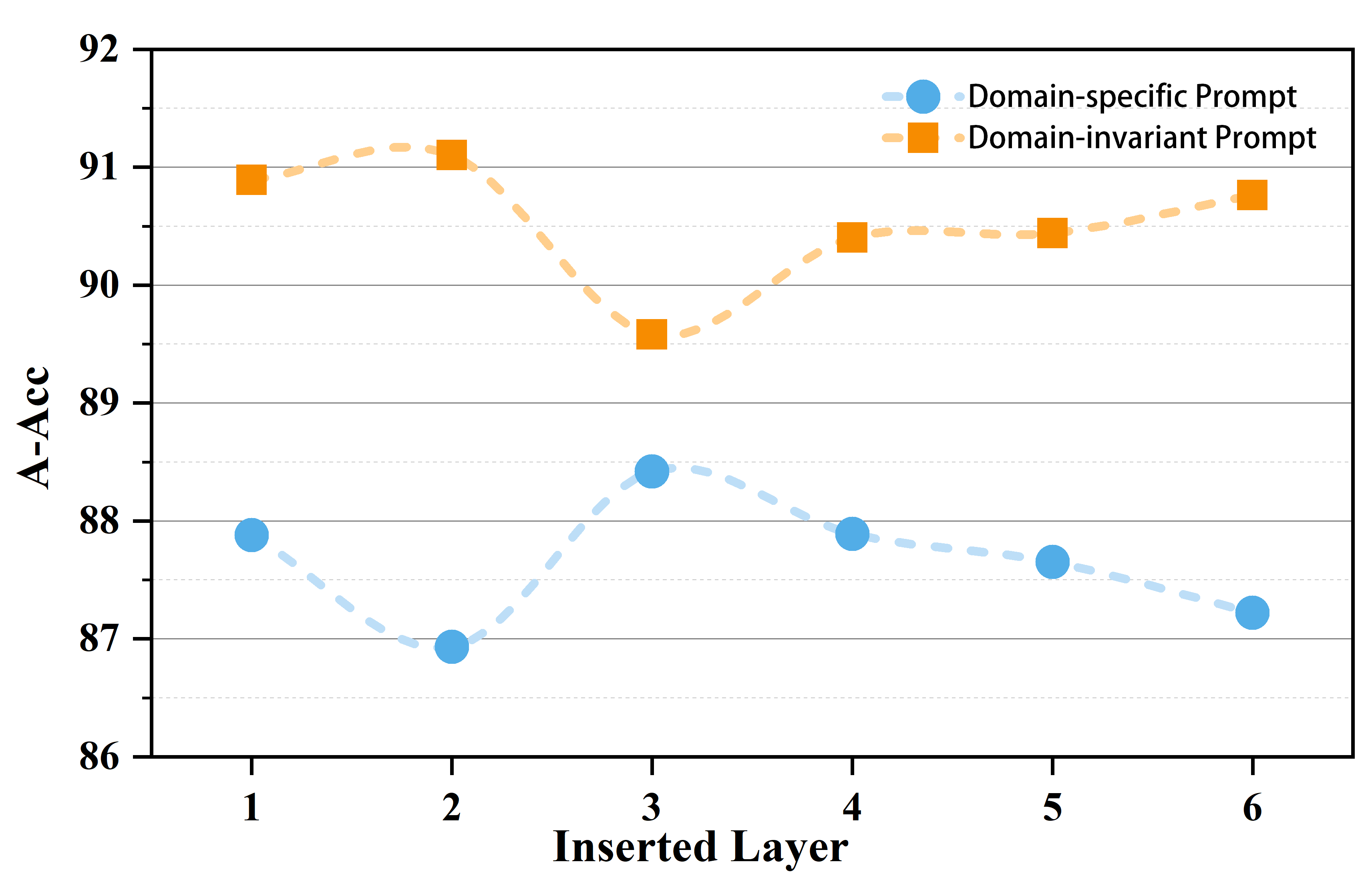}
    \caption{Searching results for the inserted layer of prompts}
  \label{fig:i-j_th}
\end{figure}

In this study, we performed a grid search on the potential positions of attaching prompts and observed the average accuracy (A-Acc) to discover the optimal positions.
According to the empirical observations of DualPrompt~\cite{wang2022dualprompt}, prompt tuning the shadow layers  generally exhibit better performance than tuning the deeper ones.
Being aware of that, we conducted a thorough grid search on the front half of layers in our backbone (\ie, from the 1st layer to the 6th layer).
We first investigated the positions of domain-specific prompts, as it has proven to be an important strategy to avoid the zero-sum game across domains~\cite{wang2022s}.
We presented the search results in Fig.~\ref{fig:i-j_th}, where the model could reach the best results and the second best results when tuning domain-specific prompts on the 3rd layer and the 4th layer respectively.
In this regard, we empirically attach the domain-specific prompts in the 3rd and 4th layers.

After determining the optimal positions of domain-specific prompts, we fixed their locations and experimented on the possible positions of domain-invariant prompts to find the optimal one.
As shown in Fig.~\ref{fig:i-j_th}, the model could achieve the best results and the second best results when attaching the domain-invariant prompts in the 2nd layer and the 1st layer respectively, suggesting they are the optimal positions for the domain-invariant prompts.

\subsubsection{Analysis on the prompt length}
In this study, we experimented with different prompt lengths to determine the optimal one.
We reported the experiment results in Table~\ref{table: sup_prompt length}.
When setting the prompt length as 5, the model reached the highest A-Acc, suggesting it is the optimal value.

\begin{table}[th!]
\centering
\caption{Searching results for the prompt length}
\begin{tabular}{c|c|c|c|c|c}
   \toprule[1pt]
    Length & 1 & 3 & 5 & 7 & 10 
\\
 \midrule
   A-Acc  & 90.24 & 91.47 & \textbf{91.61} & 91.22 & 91.25
\\
   \bottomrule[1pt]
\end{tabular}
\label{table: sup_prompt length}
\end{table}

\end{document}